\documentclass{article}
\usepackage{ijcai22}

\usepackage{times}
\usepackage{soul}
\usepackage{url}
\usepackage[hidelinks]{hyperref}
\usepackage[utf8]{inputenc}
\usepackage[small]{caption}
\usepackage{graphicx}
\usepackage{amsmath,amssymb,amsthm,mathbbol,amsfonts}
\DeclareSymbolFontAlphabet{\mathbb}{AMSb}
\DeclareSymbolFontAlphabet{\mathbbl}{bbold}
\usepackage{booktabs}
\usepackage{algorithm}
\usepackage[noend]{algpseudocode}
\usepackage{textcomp}
\usepackage{katsuki}
\usepackage{array}
\usepackage{bm}
\usepackage{nicefrac}

\usepackage{soul}
\usepackage[hidelinks]{hyperref}
\usepackage[utf8]{inputenc}
\usepackage{booktabs}
\usepackage{arydshln}
\usepackage[switch]{lineno}

\title{Cumulative Stay-time Representation for Electronic Health Records\\in Medical Event Time Prediction}

\author{
Takayuki Katsuki$^1$\and
Kohei Miyaguchi$^1$\and
Akira Koseki$^1$\and
Toshiya Iwamori$^1$\and\\
Ryosuke Yanagiya$^2$\And
Atsushi Suzuki$^2$
\affiliations
$^1$IBM Research – Tokyo, $^2$Fujita Health University
\emails
kats@jp.ibm.com, miyaguchi@ibm.com, {akoseki,iwamori}@jp.ibm.com, {yanagiya,aslapin}@fujita-hu.ac.jp
}

\algnewcommand{\Initialize}[1]{%
  \State \textbf{Initialize:} \hspace{4pt}\parbox[t]{0.7\linewidth}{\raggedright #1}
}

\algrenewcommand\algorithmicdo{}
\algrenewcommand\algorithmicindent{1.0em}

\begin{document}
\maketitle

\begin{abstract}
We address the problem of predicting when a disease will develop, i.e., medical event time~(MET), from a patient's electronic health record~(EHR).
The MET of non-communicable diseases like diabetes is highly correlated to cumulative health conditions, more specifically, how much time the patient spent with specific health conditions in the past.
The common time-series representation is indirect in extracting such information from EHR because it focuses on detailed dependencies between values in successive observations, not cumulative information.
We propose a novel data representation for EHR called cumulative stay-time representation~(CTR), which directly models such cumulative health conditions.
We derive a trainable construction of CTR based on neural networks that has the flexibility to fit the target data and scalability to handle high-dimensional EHR.
Numerical experiments using synthetic and real-world datasets demonstrate that CTR alone achieves a high prediction performance, and it enhances the performance of existing models when combined with them.
\end{abstract}

\section{Introduction}
Predicting medical events, such as disease progression, from \emph{electronic health records} (EHR) is an important task in medical and healthcare applications~\cite{tan2020data}. The EHR represents a patient's health history. Such prediction can assist in providing detailed health guidance, e.g., for early disease detection, intervention, and the allocation of limited resources in healthcare organizations~\cite{inaguma2020increasing}.

\begin{figure}[t]
    \centering
    \includegraphics[width=60mm]{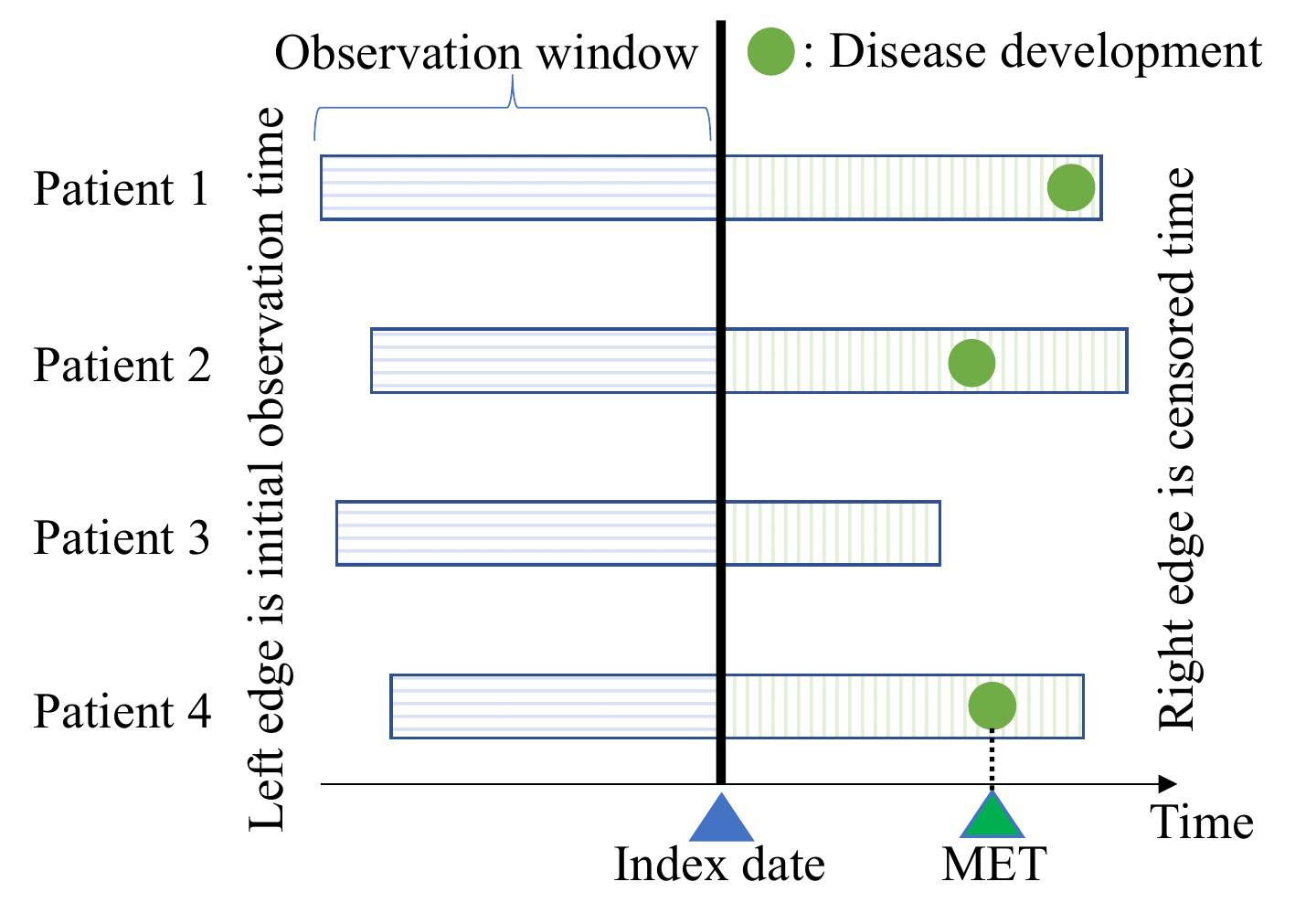}
    \caption{We predict when patient will develop disease after index date from EHR in observation window.}
    \label{FigProblem}
\end{figure}

\begin{figure}[t]
    \centering
    \includegraphics[width=80mm]{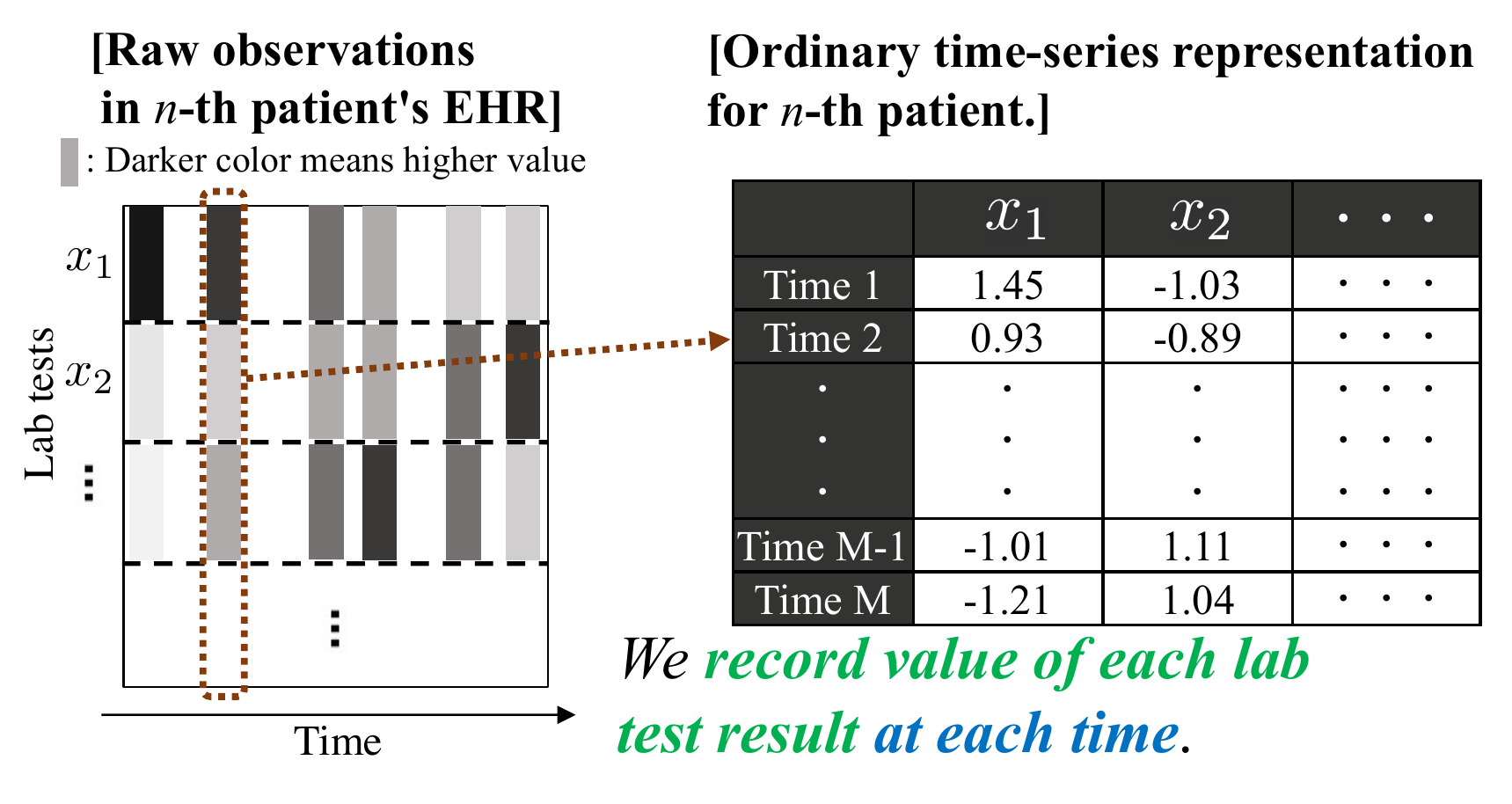}
    \caption{Ordinary time-series representation.}
    \label{FigOrdinary}
\end{figure}

This paper addresses a scenario in which we predict \emph{when} a patient will develop some disease after an index date, i.e., the \emph{medical event time}~(MET), from past observations in EHR, as shown in Fig.~\ref{FigProblem}~\cite{liu2018early}. This is a common task in survival analysis and time-to-event analysis, and we focus on MET, not just its occurrence.
The past observations for each patient come from a window that spans the initial observation time to the index date and contain lab test results at each time, as shown in the LHS in Fig.~\ref{FigOrdinary}.
From accumulated EHR datasets, we learn a prediction model for MET.



A patient's cumulative health conditions appearing in past observations in EHR are of help for MET prediction. They can be interpreted as the \emph{cumulative stay-time} in specific health states---more specifically, how much time a patient has spent with different health conditions.
For example, when a patient has high blood pressure, hyperglycemia, or high body fat for a long enough period, diseases can develop~\cite{james20142014,american20192}.
In particular, for non-communicable diseases, like diabetes,
the cumulative stay-time is extremely related to their progress and MET.

To utilize information in EHR, the common approach is to formalize the raw observations in EHR into an ordinary time-series representation~\cite{zhang2019attain,rubanova2019latent,hussain2021neural}.
In this approach, at each time, we record the value of each lab test result, as shown in the table in Fig.~\ref{FigOrdinary}. The focus is on the detailed dependencies between values in successive observations. When we handle the cumulative stay-time with this representation, prediction models, such as recurrent neural networks (RNNs)~\cite{zhang2019attain}, need to encode values in an \emph{entire time series} into the cumulative stay-time. This makes modeling the cumulative stay-time indirect.

\begin{figure}[t]
    \centering
    \includegraphics[width=84mm]{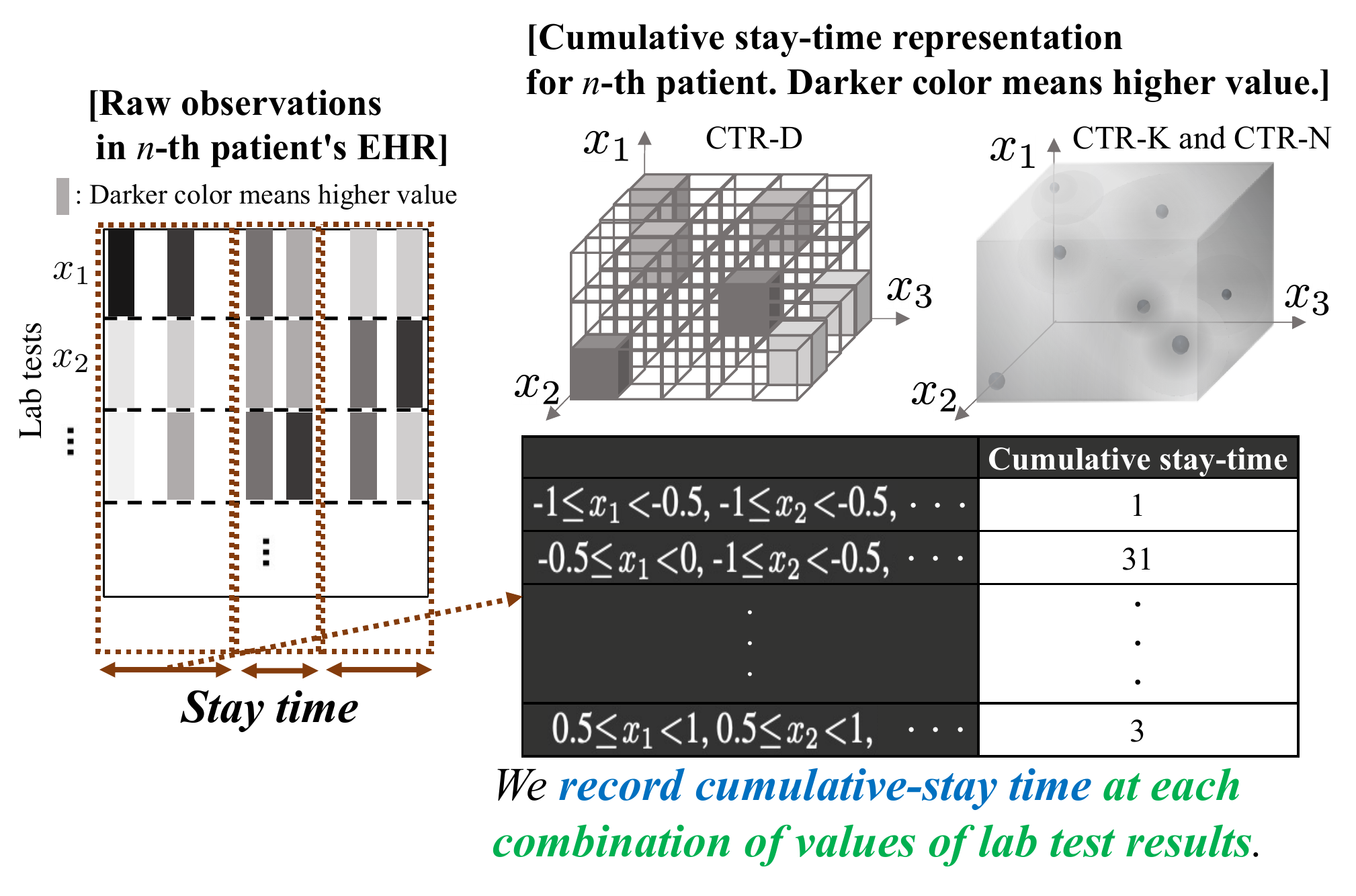}
    \caption{Cumulative stay-time representation.}
    \label{FigCumTime}
\end{figure}

We therefore propose directly extracting the cumulative stay-time from raw observations in EHR as a novel representation for EHR, that is, the \emph{cumulative stay-time representation} (CTR). In contrast to the time-series representation, we record the cumulative stay-time at each combination of values of lab test results that represents a state, as shown in Fig.~\ref{FigCumTime}. This explicitly represents how long a patient stays in a specific health state.

Representations for modeling the cumulative stay-time in specific states and using it in prediction have been proposed in other domains than EHR modeling, such as for the usage history of batteries~\cite{takahashi2012predicting} and GPS trajectories~\cite{liao2018trajectory}.
However, they are defined only with discrete state modeling that can be seen as bins of non-overlapping segmented values for lab test results, as shown in the table in Fig.~\ref{FigCumTime}. As such, they focus on low-dimensional observations, such as one, two, or three dimensions, and cannot handle more than several dimensions. This is because the number of states increases exponentially against the dimension of observation variables with this state definition. Since observations in EHR have many more dimensions, it is difficult to use these approaches on EHR directly.

This paper addresses the above difficulties by deriving methods for constructing CTR with enough scalability to handle EHR.
We first formally derive a general construction of CTR by using the discrete state. This formalization leads to further enhancements of CTR with states defined as continuous measurements, CTR-K and CTR-N, which have states based on kernel functions and neural networks, respectively.
They are more practical variants that avoid exponential increases in the number of states and lead to smooth interpolation between states.
In addition, CTR-N can be learned from data, which enables flexible state modeling.

\paragraph{Contributions.}
Our main contributions are the following:
\begin{itemize}
  \setlength{\itemsep}{0.01cm} 
  \item We propose a novel representation for EHR for MET prediction, CTR, which represents how long a patient stays in a specific health state. This helps to model the cumulative health conditions of patients.
  \item We derive a trainable construction of CTR based on neural networks that adapts to data flexibly and has scalability for high-dimensional EHR.
  \item Extensive experiments on multiple MET prediction tasks with synthetic and real-world datasets show the effectiveness of CTR, especially for EHR with relatively longer observation periods, where cumulative health conditions are more crucial for MET.
  CTR shows modularity high enough to further improve the prediction performance when combined with other models.
\end{itemize}


\section{Preliminary}
\subsection{Medical Event Time Prediction}
Our goal is to construct a model for predicting the medical event time~(MET), $y > 0$, after an index date on the basis of pairs of past observations and the corresponding timestamps, $\{\bX, \bt\}$, which are recorded in the EHR of a patient~\cite{liu2018early}, as shown in Fig.~\ref{FigProblem}. The past observations for each patient contain $M$ number of observations in an observation window $\bX \equiv \{\bx^{\{\!m\!\}}\}_{m=1}^{M}$, where the $m$-th observation $\bx^{\{\!m\!\}}$ is represented as a $D$ number of lab test results $\bx^{\{\!m\!\}}\in \mathbb{R}^{D}$, and $\bX$ thus forms an $M \times D$ matrix.
The timestamps are $\bt \equiv \{t^{\{\!m\!\}}\}_{m=1}^{M}$, where the $m$-th timestamp is $t^{\{\!m\!\}}>0$.
We assume that each patient has an $M \times D$ matrix $\bX$ and $M$ vector $\bt$.
Note that observation intervals can vary over time, and the length of sequence $M$ can be different over patients.

When we take a machine learning-based approach, the raw observations $\{\bX, \bt\}$ must be formalized into a tractable representation that contains enough information for the MET prediction.
We here denote the representation as a function, $\bz\equiv\bz(\bX, \bt)$, whose output forms either a vector, matrix, or tensor depending on the formalization.
Once $\{\bX, \bt\}$ is formalized into $\bz$, we use $\bz$ as the input of the prediction model, $f(\bz)$, and learn it with a general scheme for minimizing the expected loss:
\begin{align}
    \label{EqPredictionError}
    f^* \equiv \argmin_f E[\calL(f(\bz),y)],
\end{align}
where $f^*$ is the optimal prediction model, $\calL$ is the loss function (e.g., the squared error), and $E$ denotes the expectation over $p(y,\bX, \bt)$.
In the Experiments section, we will define a specific prediction model and loss function for each specific problem.
By using the learned $f^*$, we can predict $y$ for new data as $\hy = f^*(\bz)$.

\subsection{Property for Cumulative Stay-time Representation}
This paper focuses on how to formalize raw observations $\{\bX, \bt\}$ into a tractable representation, $\bz$.
We directly model the cumulative stay-time of a specific patient's states with the construction of $\bz$.
We would like the representation to be:
\begin{description}
    \setlength{\itemsep}{0.01cm} 
    \item[(i) Direct to model stay time]~\\
    How long a patient stays in a specific health condition can make diseases develop, particularly non-communicable diseases, like diabetes~\cite{james20142014,american20192}.
    \item[(ii) Explicit to handle variable observation intervals]~\\
    Observation intervals in EHR can vary over time and over patients since the observations are recorded when a patient is treated.
    \item[(iii) Scalable to high-dimensional data]~\\
    Since EHR has high-dimensional observations, the representation should not cause combinatorial explosion against the number of dimensions in observations.
\end{description}

\subsection{Representations for EHR}
We first discuss the conventional related representations in this subsection considering the above properties; then, we derive the cumulative stay-time representation (CTR) for EHR in Section~3.

\subsubsection{Ordinary Time-series Representation}
In the ordinary time-series representation, raw observations $\{\bX, \bt\}$ are converted into the representation of a matrix form, $\bz_{\mathrm{ts}} \in \mathbb{R}^{M \times D}$, whose two-dimensional index respectively represents timestamps and lab test names, as shown in Fig.~\ref{FigOrdinary}.
This corresponds to directly using the lab test results $\bX$ as $\bz_{\mathrm{ts}}$, $\bz_{\mathrm{ts}}(\bX, \bt) \equiv \bX$.
In the matrix, successive observations are put in adjacent rows.
Thus, such representation helps in modeling the detailed dependencies between values in successive observations by using, for example, RNNs~\cite{xiao2017modeling,zhang2019attain}, hidden Markov models (HMMs)\cite{alaa2019attentive,hussain2021neural}, convolutional neural networks (CNNs)~\cite{cheng2016risk,makino2019artificial}, ODERNNs~\cite{rubanova2019latent}, and Transformers~\cite{luo2020hitanet}.

However, when we handle the cumulative stay-time, we need to consider how long and what state a patient has been in \emph{over entire observations}, and similar health states are highly related to each other even if they are time-distant. In this case, models need to learn to encode values in an entire time series into the cumulative stay-time completely from data. The learning thus becomes indirect and costly, which leads to degraded performance.
These approaches can still handle variable observation intervals by inputting the timestamps or intervals between observations as additional inputs. Also, they usually have good scalability for high-dimensional data. Therefore, we will show the performance difference with the proposed method in our experiments to investigate their indirectness in modeling stay time in a specific health state.

\subsubsection{Cumulative Stay-time Representation}
As discussed in Introduction, methods for directly modeling the cumulative stay-time in specific states have been proposed in domains other than EHR, e.g., cumulative stay-time for the internal state of batteries~\cite{takahashi2012predicting}, location grids against GPS trajectories~\cite{andrienko2007visual,boukhechba2018predicting}, and indoor positioning using RFID~\cite{zuo2016prediction,jiang2018research}.
They can handle variable observation intervals more naturally than the time-series representation.

The details of this approach and its practical limitations in EHR modeling will be described in Section~3.1. In brief, these methods do not have enough scalability for high-dimensional data because of their discrete state modeling and have not been defined formally with further extendibility. In this paper, we formally derive this approach as the \emph{cumulative stay-time representation with discrete states} (CTR-D) and extend its state definition with a kernel function and trainable neural networks to address a higher dimensional case of EHR.

\subsubsection{Other Representations}
Representations for temporal observations other than time-series representations and CTR have been studied as reviewed in~\cite{fu2011review}, such as methods based on binarization~\cite{bagnall2006bit} and segmented subsequences~\cite{lovric2014algoritmic}.
Changing the domain from time into other domains based such as on the Fourier transform~\cite{agrawal1993efficient} and tensor factorization~\cite{ho2014marble,yu2016temporal,yin2019learning} is another common way.
These methods assume high frequency and regular observation intervals, which is not the case in our scenario.

\section{CTR: Cumulative Stay-time Representation}
We propose a cumulative stay-time representation, CTR, for directly modeling the cumulative stay-time of a specific patient's states as a novel formalization of raw observations in EHR.

\subsection{CTR-D: CTR with Discrete States}
We convert raw observations $\{\bm{X}, \bm{t}\}$ into the cumulative stay-time at a finite $K$ number of states as $K$-dimensional vector $\bz$, whose $k$-th element is $z_k > 0$.
Each state represents a combination of observed attribute values and can be seen as a bin segmented by a lattice that defines the value range of each attribute in each state, as shown in Fig.~\ref{FigCumTime}.
We cumulatively fill each bin with the stay time of which the raw observation falls into the corresponding value ranges.

By using the state function $\bs(\bm{x}^{\{\!m\!\}})\in \{0,1\}^K$, which outputs a one-hot vector representing the current state for input observation $\bm{x}^{\{\!m\!\}}$, CTR $\bz$ is defined as
\begin{align}
    \label{EqCTR}
\bm{z}(\bX, \bt) &\equiv \sum_m d^{\{\!m\!\}}\bm{s}\left(\bm{x}^{\{\!m\!\}} \right),\\\nonumber
\mathrm{where}~~&d^{\{\!m\!\}} \equiv \lambda^{t^{\{\!M\!\}}-t^{\{\!m\!\}}} (t^{\{\!m\!\}}-t^{\{m-1\}})
\end{align}
where $d^{\{\!m\!\}}$ is the stay time for the $m$-th observation, which is estimated by calculating the difference between consecutive timestamps $t^{\{\!m\!\}}$ and $t^{\{m-1\}}$ with decay for weighting newer observations. $\lambda$ is the decay rate and is optimized in training.
Since the output of the function $\bs(\bm{x}^{\{\!m\!\}})$ is a one-hot vector, only one element in the vector can become $1$, and the others are $0$, so the index for the element with value $1$ represents the current state of the patient.
Thus, for the $m$-th observation, the element in $d^{\{\!m\!\}}\bm{s}\left(\bm{x}^{\{\!m\!\}} \right)$ with the current state becomes just $d^{\{\!m\!\}}$, and the others are $0$.
Through the summation of $d^{\{\!m\!\}}\bm{s}\left(\bm{x}^{\{\!m\!\}} \right)$ over $m$, each element of $\bz$ represents the sum of the stay time in each state over the observations. Also, from Eq.~\eqref{EqCTR}, this representation can explicitly handle variable observation intervals without any additional encoding.
The algorithm is described in Algorithm~\ref{alg1}.

The state function $\bs(\bm{x}^{\{\!m\!\}})$ is defined by the indication function $\bI$, which always outputs a $K$-dimensional one-hot vector representing the current state:
\begin{equation}
    \label{EqStateDis}
\bs(\bm{x}^{\{\!m\!\}}) \equiv \bm{I}\left(\bm{x}^{\{\!m\!\}}, \bm{A} \right),
\end{equation}
where $\bm{A} \in \{\bm{a}_k \}_{k=1}^K$ is the $K$ number of non-overlapping collectively exhaustive value segments. The detailed definition of $\bm{a}_k$ and the $k$-th element of the function $\bm{I}$ are in the appendix.
If $\bx^{\{\!m\!\}}$ falls into the $k$-th segment, only the $k$-th element of $\bm{I}\left(\bm{x}^{\{\!m\!\}}, \bm{A} \right)$ becomes $1$, and the others are $0$ because of the non-overlapping segmentation.
An example segmentation is shown in the table in Fig.~\ref{FigCumTime}, which is based on equally spaced boundaries over the value range of $\bx^{\{\!m\!\}}$, [-1, -0.5, 0, 0.5, 1], where $x^{\{\!m\!\}}_d$ is defined in $[-1, 1)$. For example, in a $3$-dimensional case, $K=4^3=64$.

We call CTR in Eq.~\eqref{EqCTR} with the state function in Eq.~\eqref{EqStateDis} \emph{CTR with discrete states} (CTR-D).
The discretely defined state $\bs(\bm{x}^{\{\!m\!\}})$ is easy to understand.
When the number of attributes in $\bx$ is small enough, we can practically use the function $\bs(\bm{x}^{\{\!m\!\}})$ in Eq.~\eqref{EqStateDis} for computing $\bz$.

However, since the number of combinations representing states grows exponentially with the number of attributes $D$, CTR-D cannot handle more than a few variables.
Observations in EHR have many more attributes in general. For example, when we set the number of segments to $100$, $K$ becomes $100^{D}$, which quickly causes a combinatorial explosion according to the number of attributes $D$.
Also, the non-continuous boundary prevents generalization between adjacent states, though adjacent states should represent states similar to each other because of the shared boundaries between them in our definition in Eq.~\eqref{EqStateDis} (see also Appendix A).
We thus extend the function $\bs(\bm{x}^{\{\!m\!\}})$ into a more practical one in the following sections.
\begin{algorithm}[t]
\caption{Cumulative stay-time representation}
\label{alg1}
\begin{algorithmic}[1]
    \Require Raw observations $\{\bX, \bt\}$ and state function $\bs(\bullet)$
    \Ensure Cumulative stay-time representation, CTR, $\bz$
    \Initialize{$\bm{z} \leftarrow \bm{0}$}
    \For{$m=1$ \textbf{to} $M$~~(which can be parallelized over $m$)}
    \State $\bs^{\{\!m\!\}} \leftarrow \bs(\bm{x}^{\{\!m\!\}})$
    \State $d^{\{\!m\!\}} \leftarrow \lambda^{t^{\{\!M\!\}}-t^{\{\!m\!\}}} (t^{\{\!m\!\}}-t^{\{m-1\}})$
    \State $\bm{z} \leftarrow \bm{z} + d^{\{\!m\!\}} \bs^{\{\!m\!\}}$
    \EndFor
\end{algorithmic}
\end{algorithm}

\subsection{CTR-K: CTR with Kernel-defined States}
For mitigating the exponential growth in the number of states, we change the definition of states in Eq.~\eqref{EqCTR} from discrete, i.e., what variable values an observation has, to continuous, i.e., \emph{how close an observation is to some basis vectors}, as shown in Fig.~\ref{FigCumTime}.
Continuous states are no longer represented as a one-hot vector corresponding to a unique state; they are represented as a weight vector determining at what proportion we assign the current stay time to each state represented by bases.
In this case, the number of states is limited to the number of bases and does not grow exponentially.
This also leads to interpolation between states and can smoothly represent intermediate states between the states.

We use a kernel function that represents affinities to bases for observations, where we construct the continuous state vector by assigning different values to multiple elements according to the affinities.
The state function $\bs_{\mathrm{K}}(\bm{x}^{\{\!m\!\}})\in \mathbb{R}^K$ based on the kernel function $\bm{\phi}$ is defined as
\begin{equation}
    \label{EqStateCon}
\bs_{\mathrm{K}}(\bm{x}^{\{\!m\!\}}) \equiv \bm{\phi}\left(\bm{x}^{\{\!m\!\}},\bm{B} \right),
\end{equation}
where $\bm{B} \equiv \{ \bm{b}^{\{\!k\!\}} \}_{k=1}^K$ is the $K$ number of bases, and $\bm{b}^{\{\!k\!\}} \in \mathbb{R}^D$ is the $k$-th basis.
For example, $\bs_{\mathrm{K}}(\bm{x}^{\{\!m\!\}}) = \{0,0.3,0.7,0,...,0 \}$ means that we assign the stay time for the $m$-th observation with weights of $0.3$ and $0.7$ to the second and third states, respectively, in the summation in Eq.~\eqref{EqCTR}. Bases can be randomly sampled from the training set.
When the observation variables are real-valued, as in our scenario, the choice of $\bphi$ is an RBF kernel, whose definition is provided in Appendix B. We can also use other kernels, such as tf-idf vector $+$ cosine similarity~\cite{rajaraman2011mining}, for binary features.

We call CTR in Eq.~\eqref{EqCTR} with the state function in Eq.~\eqref{EqStateCon} \emph{CTR with kernel-defined states} (CTR-K).


\subsection{CTR-N: CTR with Neural Network-defined States}
Additionally, we can consider the requirement for continuous state $\bs_{\mathrm{K}}(\bm{x}^{\{\!m\!\}})$ in Eq.~\eqref{EqStateCon} to represent a similar observation with a similar weight vector.
Such a vector can also be modeled with neural networks since they are trained to produce similar outputs from similar inputs.

We thus extend $\bs_{\mathrm{K}}(\bm{x}^{\{\!m\!\}})$ to $\bs_{\mathrm{N}}(\bm{x}^{\{\!m\!\}})\in \mathbb{R}^K$ by replacing kernel function $\bphi$ with a trainable neural network, $\bg$, e.g., multilayer perceptron (MLP), that produces a state-indicating weight vector similar to $\bm{\phi}$, as
\begin{align}
    \label{EqStateNeural}
\bs_{\mathrm{N}}(\bm{x}^{\{\!m\!\}}) &\equiv \bg\left(\bm{x}^{\{\!m\!\}},\bm{\theta}_g \right),
\end{align}
where $\bm{\theta}_g$ are parameters for the neural network. The outputs of the final layer for $\bg$ should be normalized, such as by the softmax function, as a weight vector, and the number of the outputs is $K$.
The specific neural network structure for $\bg$ is shown in the Experiments section.

We call CTR in Eq.~\eqref{EqCTR} with the state function in Eq.~\eqref{EqStateNeural} \emph{CTR with neural network-defined states} (CTR-N).
This representation can be learned from data and thus provides more flexibility in adjusting the state definition to target data. Also, in contrast to CTR-D and CTR-K, CTR-N does not require having to choose the state boundaries or the bases.

\newtheorem{lemma}{Lemma}
Formally, the following lemma characterizes CTR-K and CTR-N with respect to the three properties: (i) direct to model stay time, (ii) explicit to handle variable observation intervals, and (iii) scalable to high-dimensional data:
\begin{lemma}
  (i) Every element in $\bm{z}(\bX, \bt)$ is a linear function of stay time $d$. Hence, $\bm{z}(\bX, \bt)$ is a direct representation of stay time.
  (ii) $\bm{z}(\bX, \bt)$ is a function of an observation interval $t^{\{\!m\!\}}-t^{\{m-1\}}$.
  (iii) The number of dimensions in $\bs_{\mathrm{K}}(\bm{x})$ and that of the corresponding $\bm{z}(\bX, \bt)$ depend on the number of bases, not the number of attributes in $\bx$, $D$. Also, the number of dimensions in $\bs_{\mathrm{N}}(\bm{x})$ and that of the corresponding $\bm{z}(\bX, \bt)$ depend on the number of outputs of $\bg$, not $D$.
\end{lemma}
\noindent The ordinary time-series representation $\bz_{\mathrm{ts}}(\bX, \bt)$ does not satisfy (i). CTR-D does not satisfy (iii).

\paragraph{Gradients for learning model parameters.}
We minimize Eq.~\eqref{EqPredictionError} by using gradient-based optimization methods.
We learn $f$ on the basis of Eqs.~\eqref{EqPredictionError} and~\eqref{EqCTR} by using the gradients for the model parameters for $f$, $\btheta_f$, as
\begin{align}
    \label{gradient_f}
    \frac{\partial \calL}{\partial \btheta_f} =& E \bigg[\frac{\partial \calL} {\partial f} \frac{\partial f} {\partial \btheta_f}\bigg],
\end{align}
where we omit the inputs of the functions for simplicity.

For CTR-N, in addition to learning $f$, we learn the parameters $\btheta_g$ of neural network $\bg$ in~\eqref{EqStateNeural}, which represents the state $\bs_{\mathrm{N}}(\bm{x}^{\{\!m\!\}})$.
The gradients for $\btheta_g$ can be derived as
\begin{align}
    \label{gradient_N}
    \frac{\partial \calL}{\partial \btheta_g} =& E \bigg[\frac{\partial \calL} {\partial f} \frac{\partial f} {\partial \bz} \sum_m d^{\{\!m\!\}} \frac{\partial g\left(\bm{x}^{\{\!m\!\}},\bm{\theta}_g \right)} {\partial \btheta_g}\bigg].
\end{align}

\section{Experiments}
\begin{figure}[t]
    \centering
    \includegraphics[width=84mm]{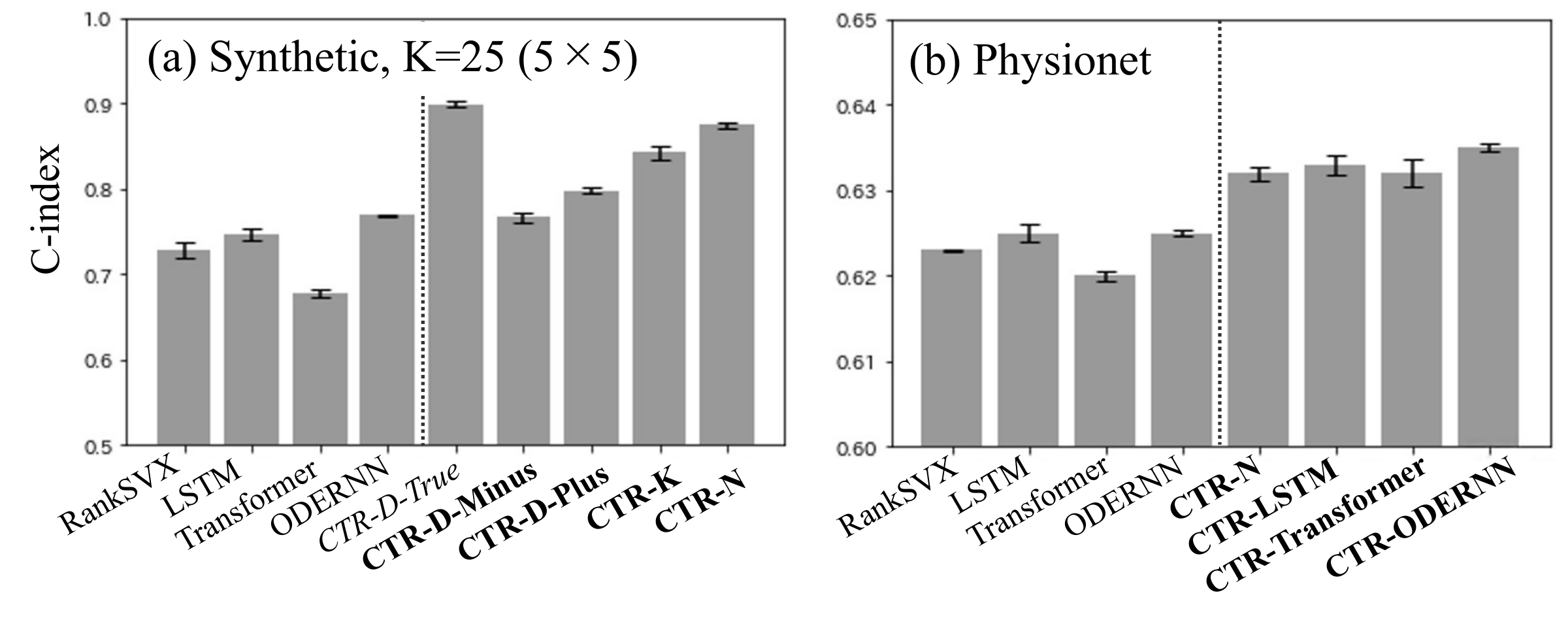}
    \caption{Comparison of C-index on (a) Synthetic and (b) Physionet (higher is better).}
    \label{FigExSynPhy}
\end{figure}
We assessed the prediction performance of our method CTR in numerical experiments to show the effectiveness of directly modeling the cumulative stay-time.

\paragraph{Evaluation metric.}
We report the mean and standard error of the concordance index (C-index)~\cite{liu2018early} across $5$-fold cross-validation, each with a different randomly sampled training-testing split. The C-index becomes a high value when regressed MET values follow the true ordering, and it can handle censored data, which is essential for our application with EHR.
For each fold of the cross-validation, we randomly sampled $20\%$ of the training set for use as a validation set to determine the best hyperparameters for each method, where hyperparameters providing the highest C-index in the validation set were chosen.

\paragraph{Implementations for CTRs.}
We used the loss proposed in~\cite{liu2018early}, which can handle censored data well.
We used a $3$-layer MLP with ReLU~\cite{nair2010rectified} (more specifically, $D$-$100$-$1$) as the prediction model $f$ for CTR-D, CTR-K, and CTR-N. In CTR-N, we used a $4$-layer MLP with ReLU (more specifically, $D$-$100$-$100$-$100$) as $\bg$ in Eq.~\eqref{EqStateNeural}, where the final layer is the softmax function for normalization.
More details on the implementation, such as the definitions for states in CTR-D and CTR-K, are provided in the appendix.
Note that we could not use CTR-D for real-world EHR experiments since we needed to handle a large number of attributes, which would cause a combinatorial explosion for $K$. For example, when $D=37$ and we set the number of segments to $100$, $K$ becomes $100^{37}$, where one of our real-world datasets contains $D=37$ number of attributes.

\paragraph{Methods compared.}
The proposed method was compared with four state-of-the-art methods:
\textbf{RankSVX}~\cite{liu2018early}, \textbf{LSTM}, \textbf{Transformer}~\cite{luo2020hitanet}, and \textbf{ODERNN}~\cite{rubanova2019latent}, where the loss for each method is the same as CTRs, and the number of hidden states in each model is the same as CTR-N.
They are based on the time-series representation with the stay time $\bd$ treated as another column for the input time series. Details on their implementation are provided in the appendix.

\paragraph{Combinations of CTR-N and compared methods.}
In real-world EHR experiments, we also examined combinations of CTR-N with the compared methods LSTM, Transformer, and ODERNN (\textbf{CTR+LSTM}, \textbf{CTR+Transformer}, and \textbf{CTR+ODERNN}, respectively). In these combinations, the representation just before the final linear layer of each model was extracted, concatenated with CTR $\bz$ as a single vector, and fed into the prediction model $f(\bz)$. They were trained in an end-to-end manner.

\paragraph{Computing infrastructure.}
All of the experiments were carried out
on workstations having $128$ GB of memory, a $4.0$-GHz CPU, and an Nvidia Tesla V100 GPU.
The computational time for each method was a few hours for producing the results for each dataset, \emph{except for ODERNN, which was $10$~to~$20$~times slower than the other methods.}

\subsection{Results}
We first use a synthetic dataset to investigate \emph{1) whether the method with CTR can indeed learn to predict what cannot be learned without CTR.}
Then, real-world EHR datasets are used to show the \emph{2) practical effectiveness of CTR.}
Finally, we show that \emph{3) CTR enhances the prediction performance, especially for EHR with relatively longer observation periods}, where cumulative health conditions are more crucial for MET. Details on the datasets are provided in the appendix.

\paragraph{Synthetic.}
The Synthetic dataset was generated on the basis of our assumed nature, i.e., the cumulative stay-time for each state leads to the development of a disease.
The number of records was $N=1,000$, the observation length for each record was $M=10$, and the number of attributes was $D=2$. The observation intervals varied between records.
We addressed large ($K=100$), medium ($K=49$), and small ($K=25$) numbers of states settings in data generation.
The results are shown in Fig.~\ref{FigExSynPhy}-(a), where the bars represent the means of the C-index across $5$-fold cross-validation, and the confidence intervals are standard errors. We show the results with the small ($K=25$) number of states here, and the others are provided in the appendix.
We can see that the overall performance of the proposed method was significantly better than those of the compared methods, which demonstrates that \emph{the proposed method with CTR can learn what cannot be learned without CTR well.}
Note that we used multiple settings for CTR-D: the same number of states $K$ for the data generation (CTR-D-True), $K_d-1$ (CTR-D-Minus), and $K_d+1$ (CTR-D-Plus), where $K_d \equiv \sqrt[2]{K}$. We used CTR-D-True as the reference when we knew the true CTR; it thus should achieve the highest score. CTR-K and CTR-N were better than CTR-D with the wrong number of states even if the error was $1$, which demonstrates that CTR-K and CTR-N have a better generalization capability than CTR-D against data variation.
CTR-N performed the best, which demonstrates that CTR-N learns states from data well.

\paragraph{Physionet.}
The Physionet dataset is a publicly available real-world EHR dataset (Physionet Challenge 2012~\cite{silva2012predicting}). The number of records was $N=8,000$, and the number of attributes was $D=37$. The observation intervals varied between records.
The results for the MET prediction task for patient death are shown in Fig.~\ref{FigExSynPhy}-(b) with the same configuration as the results of the Synthetic dataset.
The performances of the methods with CTR were better than those of the methods without CTR by a sufficient margin in terms of standard error.
These results demonstrate that CTR can improve the C-index in the MET prediction problem with real-world EHR. We omitted the results with CTR-K since it was always worse than CTR-N.
CTR-N achieved the best performance on average in comparison with the single models. In addition, when looking at results for combinations of CTR-N and other models, CTR+LSTM, CTR+Transformer, and CTR+ODERNN, we can see that adding CTR-N to these models improved their performance further, which shows the high modularity of CTR to work complementarily with other models.
\emph{This shows that CTR and the time-series models captured different temporal natures in real-world EHR.}
We can automatically determine which type of temporal natures to take into account with the training dataset by training the prediction model $f$, which is put on top of these models.

\paragraph{Case study.}
The above experiments on two different datasets have shown that the methods with CTR have superior prediction performance compared with the state-of-the-art methods for MET prediction from EHR.
Here, we show a real healthcare use-case, where we predict the onset of complications with diabetes mellitus from a real-world big EHR database.
We used datasets provided by one of the largest hospitals in Japan that has maintained a big database of more than $400,000$ patients since $2004$~\cite{makino2019artificial,inaguma2020increasing}.
We worked with six datasets for six kinds of complications of diabetes mellitus: hyperosmolar (HYP), nephrology (NEP), retinopathy (RET), neuropathy (NEU), vascular disease (VAS), and other complications (OTH), each of which has over $N=15,000$ records. The number of attributes was $D=26$, and the observation intervals and lengths varied between records.
In this scenario, ODERNN, which is $10$ to $20$ times slower than the other methods, did not meet the practical needs for this large-scale dataset.
Thus, we here show a comparison between the proposed method and the second-best baseline, LSTM, in experiments with the Synthetic and Physionet datasets.
The results of the mean and standard error of the C-index across $5$-fold cross-validation are listed in Table~\ref{tabDiabetesResults}.
For most of the six tasks having over $15,000$ samples each, the performances of the methods with CTR were better than LSTM by a sufficient margin in terms of standard error.
Complications of diabetes mellitus are known to develop due to time-cumulative effects for vessels with an unhealthy status. The results showed that our explicit mechanisms are essential to learning such effects to achieve higher prediction performance.

\begin{figure}[t]
    \centering
    \includegraphics[width=84mm]{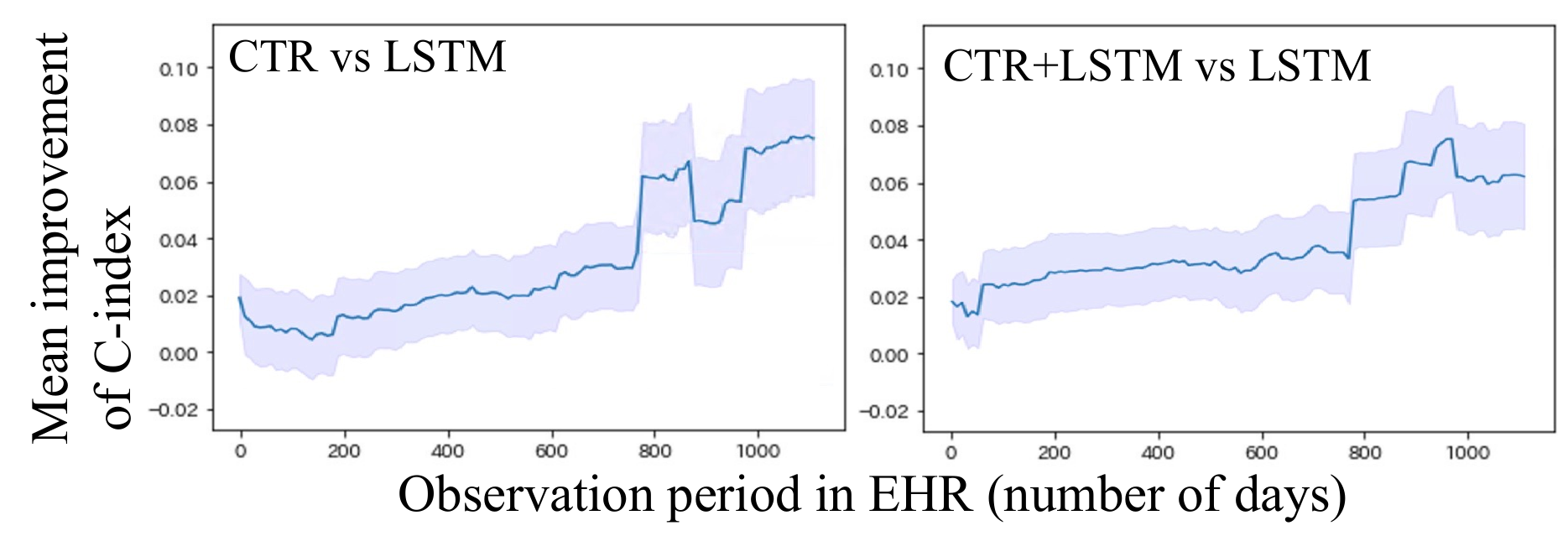}
    \caption{Analysis of performance improvement over different observation periods (higher is better).}
    \label{FigExDiffPeriod}
\end{figure}

\paragraph{Performance analysis on different observation periods.}
We further analyzed the performance improvements of the methods with CTR compared with LSTM over different observation periods by using the Case study dataset containing EHR with more extended periods.
We plotted the mean improvements of the C-index between them for data with different observation periods, as shown in Fig.~\ref{FigExDiffPeriod}, where the confidence intervals are standard errors of the improvements.
The right region in the figure show the results for data with longer observation periods.
It shows that \emph{CTR improved the performance, especially for data with relatively longer observation periods}, where cumulative health conditions are more crucial for MET prediction.
\begin{table}[t]
\centering
\small
\begin{tabular}{cccc}
\toprule
{}&\!LSTM\!&\!\textbf{CTR-N}\!&\!\textbf{CTR+LSTM}\!\\
\midrule
HYP&\!0.589$\pm$0.026\!&\!$\bm{0.612}$$\pm$0.026\!&\!0.583$\pm$0.036\!\\
NEP&\!0.689$\pm$0.012\!&\!$\bm{0.739}$$\pm$0.010\!&\!0.708$\pm$0.007\!\\
RET&\!0.717$\pm$0.013\!&\!0.721$\pm$0.026\!&\!$\bm{0.745}$$\pm$0.008\!\\
NEU&\!0.569$\pm$0.023\!&\!$\bm{0.608}$$\pm$0.020\!&\!0.600$\pm$0.020\!\\
VAS&\!0.503$\pm$0.035\!&\!0.481$\pm$0.013\!&\!$\bm{0.534}$$\pm$0.022\!\\
OTH&\!0.718$\pm$0.015\!&\!$\bm{0.741}$$\pm$0.020\!&\!0.734$\pm$0.011\!\\
\midrule
Average&\!0.664\!&\!$0.687$\!&\!$\bm{0.688}$\!\\
\bottomrule
\end{tabular}
\caption{Comparison of C-index in case study (higher is better). Confidence intervals are standard errors. Best results are in bold.}
\label{tabDiabetesResults}
\end{table}

\section{Conclusion}
We proposed a cumulative stay-time representation, CTR, for a novel representation of EHR.
CTR can efficiently handle the situation in which the development of some disease is related to the cumulative stay-time of a specific patient's health conditions, e.g., non-communicable diseases.
We developed three variations of CTR with discrete states, continuous states with kernel functions, and continuous states with neural networks.
In particular, CTR with neural networks, CTR-N, is practical because it has scalability handling high-dimensional data and can be learned from data for flexibility in adjusting to the target data.
An experimental evaluation demonstrated that the method with CTR performed better than the methods without CTR.
Application to domains other than EHR will be an interesting avenue for future work.

\fontsize{9.0pt}{10.0pt} \selectfont
\bibliographystyle{named}
\bibliography{ref}

\appendix
\section{Detailed Definition of States in CTR-D}
The state function $\bs(\bm{x}^{\{\!m\!\}})$ is defined by the indication function $\bI$, which always outputs a $K$-dimensional one-hot vector representing the current state:
\begin{equation}
    \label{EqStateDis2}
\bs(\bm{x}^{\{\!m\!\}}) \equiv \bm{I}\left(\bm{x}^{\{\!m\!\}}, \bm{A} \right),
\end{equation}
where $\bm{A} \in \{\bm{a}_k \}_{k=1}^K$ is the $K$ number of non-overlapping collectively exhaustive value segments.

The segment for the $k$-th state, $\bm{a}_k\equiv \left\{\left[\zeta_{d,k}, \xi_{d,k} \right)\right\}_{d=1}^D$, represents the combination of $D$ number of value ranges, where $\zeta_{d,k}$ and $\xi_{d,k}$ respectively represent lower and higher boundaries for the $d$-th attribute $x^{\{\!m\!\}}_d$.
By using $\zeta_{d,k}$ and $\xi_{d,k}$, the $k$-the element of the function $\bm{I}$ is
\begin{equation}
    \label{EqRule}
    \left[\bm{I}\left(\bm{x}^{\{\!m\!\}}, \bm{A} \right)\right]_k \equiv \prod_d \mathbbl{1}\left(\zeta_{d,k} \le x^{\{\!m\!\}}_d < \xi_{d,k} \right),
\end{equation}
where $\mathbbl{1}(\bullet)$ is an indication function that returns only a value of $1$ when the $\bullet$ condition is satisfied and otherwise returns $0$.
If $\bx^{\{\!m\!\}}$ falls into the $k$-th segment, only the $k$-th element of $\bm{I}\left(\bm{x}^{\{\!m\!\}}, \bm{A} \right)$ becomes $1$ and the others $0$ because of the non-overlapping segmentation.
An example segmentation is shown in the table in Fig.~2 in the main text, which is based on equally spaced boundaries over the value range of $\bx^{\{\!m\!\}}$, [-1, -0.5, 0, 0.5, 1], where $x^{\{\!m\!\}}_d$ is defined in $[-1, 1)$.
For example, in a $3$-dimensional case, $K=4^3=64$.

\section{Detailed Definition of Kernels in CTR-K}
When the observation variables are real-valued, as in our scenario, the choice of $\bphi$ is an RBF kernel defined as
\begin{equation}
    \label{EqStateKernel}
\bm{\phi}\left(\bm{x}^{\{\!m\!\}},\bm{B} \right)\equiv \left\{ \frac{\exp(-\gamma \|\bm{x}^{\{\!m\!\}}- \bm{b}^{\{\!k\!\}}\|^2)}{Z_m} \right\}_{k=1}^K,
\end{equation}
where $\gamma$ is a bandwidth parameter to be optimized with a grid search using a validation set in training data, and $Z_m\equiv \sum_k \exp(-\gamma \|\bm{x}^{\{\!m\!\}}- \bm{b}^{\{\!k\!\}}\|^2)$ is a normalizing factor for the $m$-th observation, which comes from the requirement for using $\bs_{\mathrm{K}}$ as weights for assigning the stay time in Eq. (2) in the main text.

We can also use other kernels, such as tf-idf vector $+$ cosine similarity~\cite{rajaraman2011mining}, for binary features. It is defined as
\begin{equation}
    \label{EqStateKernelBinary}
\bm{\phi}\left(\bm{x}^{\{\!m\!\}},\bm{B} \right)\equiv \left\{ \frac{\bm{\xi}(\bm{x}^{\{\!m\!\}}) \bm{\xi}(\bm{b}^{\{\!k\!\}}))}{|\bm{\xi}(\bm{x}^{\{\!m\!\}})||\bm{\xi}(\bm{b}^{\{\!k\!\}}))|} \right\}_{k=1}^K,
\end{equation}\
where $\bm{\xi}(\bullet)$ represents tf-idf function~\cite{rajaraman2011mining}.

\section{Details on Experiments with Synthetic Dataset}
We randomly generated $N=1,000$ samples, where the $n$-th sample was represented by $\{\bX^{\{\!n\!\}}, \bd^{\{\!n\!\}}\}$. Note that we directly generated duration $\bd$ instead of timestamp $\bt$. Each sample contained $M=10$ number of raw observations, and the $m$-th observation in the $n$-th sample, $\bx^{\{n,m\}}$, was generated from uniform distribution $\UniformDist(\bx^{\{n,m\}} | -1, 1)$. We set the number of attributes in $\bx^{\{n,m\}}$ to $D=2$. The corresponding duration $d^{\{n,m\}}$ was then generated from $\UniformDist(d^{\{n,m\}} | 0, 1)$.

\begin{figure}[t]
    \centering
    \includegraphics[width=80mm]{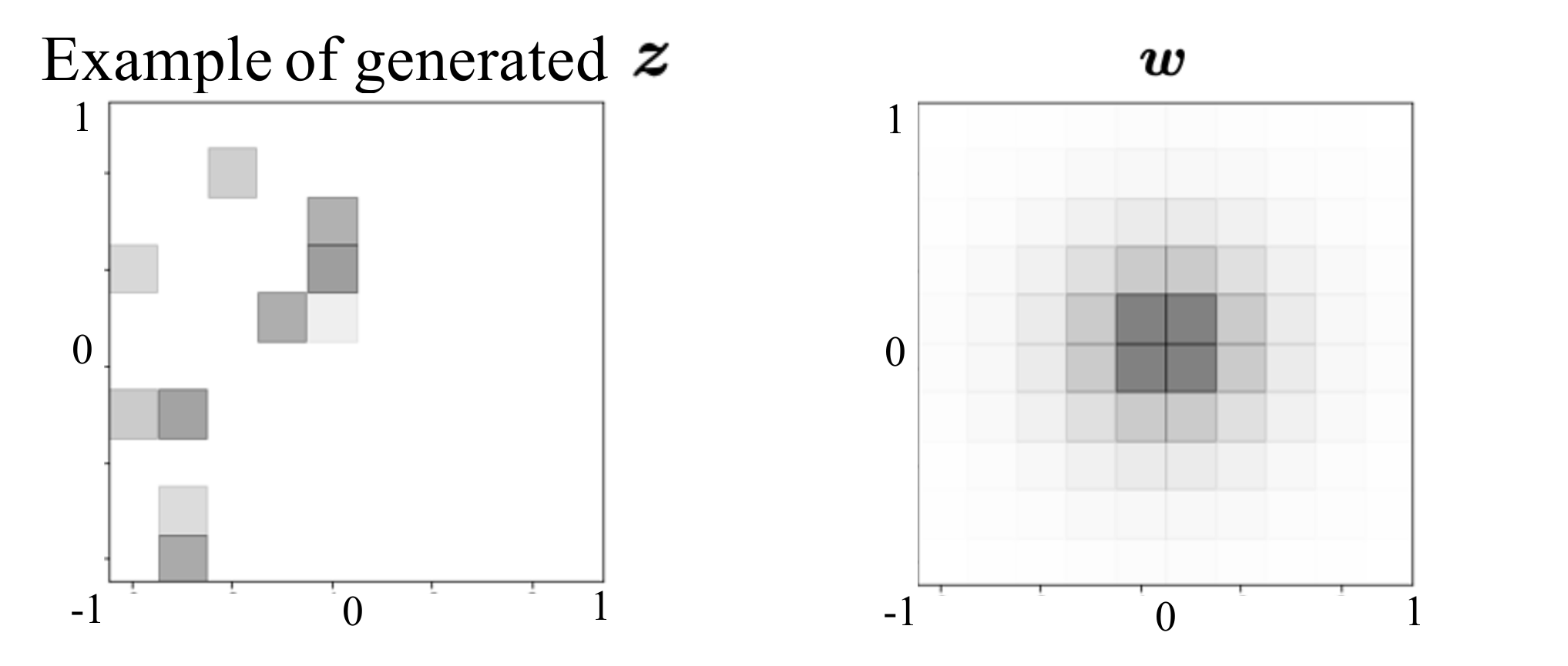}
    \caption{Examples of generated $\bz$ and $\bw$ in experiments on synthetic data ($K=100$). Darker color means higher value.}
    \label{FigExSynthetic_example}
\end{figure}

After that, using $\{\bX^{\{\!n\!\}}, \bd^{\{\!n\!\}}\}$, $\bz^{\{\!n\!\}}$ was computed by using Eqs.~(2)--(4) in the main text with equally spaced state boundaries over the value range of $\bx^{\{n,m\}}$ under different settings of numbers of states $K = \{25, 49, 100\}$. Under this $D=2$ setting, $\bz^{\{\!n\!\}}$ can be viewed as a matrix, as shown in Fig.~\ref{FigExSynthetic_example} ($K = 100$).

Finally, we generated $N$ sets of true labels $\by = \{y^{\{\!n\!\}}\}_{n=1}^{N}$ from Gaussian distribution $\NormalDist(y^{\{\!n\!\}} |\tr(\bw^\top \bz^{\{\!n\!\}}), 0.1)$, where $\bw$ is the same size of matrix as $\bz$ that was generated using a Gaussian function, whose center is the center of the bins for $\bz$ and whose width is $1$, as shown in Fig.~\ref{FigExSynthetic_example}; $\tr(\bullet)$ represents the trace of the matrix $\bullet$, and $\top$ denotes the transpose.
Since $\bw$ is generated from a smooth function over states, we have a chance to generalize among states even with a small number of samples.
Since the generation process for $y$ is non-linear and there is no longer a correlation between $y$ and $\{\bX, \bd\}$, learning with this data is difficult in general.
\begin{figure*}[t]
    \centering
    \includegraphics[width=175mm]{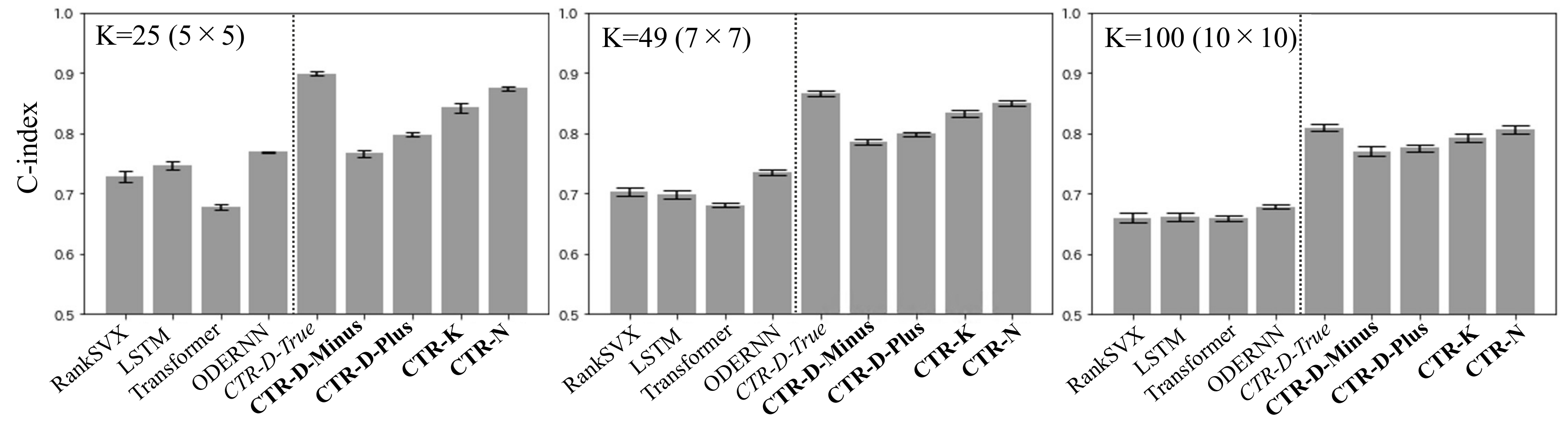}
    \caption{Comparison of C-index on synthetic data (higher is better).}
    \label{FigExSynthetic_results}
\end{figure*}

We repeatedly evaluated the proposed method for each of the following settings for the number of states $K$ in $\bz$: $K = \{25, 49, 100\}$, which are referred to as large, medium, and small numbers of state settings in the main text.
This corresponds to dividing each attribute value into $K_d \equiv \sqrt[2]{K} = \{5, 7, 10\}$ segments in this $D=2$ setting, respectively.
We trained models with only $\bX$, $\bd$, and $y$ without $\bz$.

\paragraph{Implementation.}
We used the squared loss for the loss function $\calL$ in Eq.~(1) in the main text.
Then, using training samples, we empirically estimated the expected loss $E[\calL]$ in Eq.~(1) in the main text as
\begin{equation}
    \label{EqLossSynthetic}
    E[\calL(f(\bz),y)]\simeq \frac{1}{N_{\mathrm{train}}}\sum_n ( y^{\{\!n\!\}}- f(\bz^{\{\!n\!\}})))^2,
\end{equation}
where $\bz^{\{\!n\!\}}$ is $\bz(\bX^{\{\!n\!\}},\bd^{\{\!n\!\}})$ with input $\bd^{\{\!n\!\}}$ instead of $\bt^{\{\!n\!\}}$, and $N_{\mathrm{train}}$ is the number of training samples.
We used a $3$-layer MLP with ReLU~\cite{nair2010rectified} (more specifically, $D$-$100$-$1$) as the prediction model $f$ for CTR-D, CTR-K, and CTR-N. Each specific implementation for the stats in CTR-D, CTR-K, and CTR-N was as follows.
\textbf{CTR-D:} We used equally spaced state boundaries with multiple settings of $K'$: the same value as $K$ for the data generation (CTR-D-True), $K_d-1$ (CTR-D-Minus), and $K_d+1$ (CTR-D-Plus).
Note that we used CTR-D-True as the reference when we knew the true CTR; it thus should achieve the highest score.
We here examined how closely the other methods performed to CTR-D-True.
\textbf{CTR-K:} We used the RBF kernel in Eq.~(5) in the main text as $\bphi$ in Eq.~(4) in the main text with $K=100$ number of bases randomly sampled from the training set and with the candidates of hyperparameter $\gamma$ as $\{10^{-2},10^{-1},10^{0},10^{1},10^{2} \}$.
\textbf{CTR-N:} We used a $4$-layer MLP with ReLU (more specifically, $D$-$100$-$100$-$100$) as $\bg$ in Eq.~(6) in the main text, where the final layer is the softmax function for normalization.
For optimization, we used Adam with the recommended hyperparameters~\cite{kingma2014adam}, and the number of samples in the mini batches was $64$. We also used a dropout~\cite{srivastava2014dropout} with a rate of $50\%$ and batch normalization~\cite{ioffe2015batch} after each fully connected layer for both MLPs in $f$ and $\bg$.
By using the learned $\hf$ and $\hat{\bg}$, we estimated $\hy = \hf(\bz(\bX,\bd))$ for the new data.

\paragraph{Methods for comparison.}
The proposed method was compared with four state-of-the-art methods.
\textbf{RankSVX} is the method proposed in~\cite{liu2018early} that uses the loss in Eq.~\eqref{EqLossEHR} and has the same prediction model $f$ as our method without CTR, i.e., $f(\bx_{\mathrm{dem}})$. For its input, we used mean, standard deviation, and $\{0.1, 0.25, 0.5, 0.75, 0.9\}$ quantiles in the input time series having the stay time $\bd$ treated as another column for the input time series.
\textbf{LSTM}, \textbf{Transformer}, and \textbf{ODERNN} have the same prediction model $f$ as our method but have outputs of LSTM, Transformer, and ODERNN as inputs for $f$ instead of $\bz$, i.e., $f(\bh_{\mathrm{LSTM}}(\bz_\mathrm{TS}),\bx_{\mathrm{dem}})$, $f(\bh_{\mathrm{Transformer}}(\bz_\mathrm{TS}),\bx_{\mathrm{dem}})$, and $f(\bh_{\mathrm{ODERNN}}(\bz_\mathrm{TS}),\bx_{\mathrm{dem}})$, respectively, where the number of hidden states in each model is the same as CTR-N, and $\bz_\mathrm{TS}$ is a time-series representation with stay time.
They are based on the time-series representation.

\paragraph{Results.}
The results are shown in Fig.~\ref{FigExSynthetic_results}, where the bars represent the means of the C-index across $5$-fold cross-validation, and the confidence intervals are standard errors.
We can see that the overall performance of the proposed method was significantly better than those of the compared methods, which demonstrates that the proposed method with CTR can learn what cannot be learned without CTR well.
Note that CTR-K and CTR-N were better than CTR-D with the wrong number of states even if the error was $1$, which demonstrates that CTR-K and CTR-N have a better generalization capability than CTR-D against data variation.
We also found that CTR-N performed the best, which demonstrates that CTR-N learns states for CTR from data well.

\section{Details on Experiments with Physionet Dataset}
We applied our CTR to a publicly available real-world EHR dataset (Physionet Challenge 2012~\cite{silva2012predicting}). The dataset consists of $N=8,000$ patients' records, and each record consists of $D=37$ lab test results, such as Albumin, heart-rate, glucose etc., from the first 48 hours after admission to an \emph{intensive care unit} (ICU). Details on the lab tests are at \url{https://physionet.org/content/challenge-2012/1.0.0/}.
We used this dataset for the MET prediction task for patient death.

The input EHR for the $n$-th patient was represented by $\{\bX^{\{\!n\!\}}, \bt^{\{\!n\!\}}\}$, where the raw observations $\bX^{\{\!n\!\}}$ were real-valued results of lab tests whose $m$-th observation was $\bx^{\{n,m\}}$, and the observation length for $\bX^{\{\!n\!\}}$ was $M_n$, which differed over patients.
$\bt^{\{\!n\!\}}$ represents the corresponding timestamps. The observation intervals between them vary over time. Thus, the stay time information and the direct modeling of it as a variable is crucial for prediction with this dataset.
The MET label for the $n$-th input EHR was $y^{\{\!n\!\}} > 0$, which is minutes over which a patient died.
All features were z-normalized with the exception of binary features.

\paragraph{Implementation.}
In this experiment, since there were censored data, we used the loss of a combination of both regression and ranking of MET values for $\calL$ in Eq.~(1) in the main text, which follows the work of~\cite{liu2018early}.
The ranking loss handles censored data well since it does not require a value for label $y$ but instead gets penalized if the values predicted for patients who have developed a complication earlier are larger than those predicted for patients who developed a disease or whose data were censored.
We then used training samples to empirically estimate the expected loss $E[\calL]$ in Eq.~(1) in the main text as
\begin{align}
    \label{EqLossEHR}
    &E[\calL(f(\bz),y)] \simeq \frac{1}{N_{\mathrm{train}}}\sum_n ( y^{\{\!n\!\}}- f(\bz^{\{\!n\!\}},\bx_{\mathrm{dem}}^{\{\!n\!\}}))^2 \nonumber\\
    &- \frac{1}{N_c} \sum_{\{n,l\} \in y^{\{\!n\!\}}<y^{\{\!l\!\}}} \ln \sigma(f(\bz^{\{\!n\!\}},\bx_{\mathrm{dem}}^{\{\!n\!\}})-f(\bz^{\{\!l\!\}},\bx_{\mathrm{dem}}^{\{\!l\!\}})),
\end{align}
where $\bz^{\{\!\bullet\!\}}$ is $\bz(\bX^{\{\!\bullet\!\}},\bt^{\{\!\bullet\!\}})$, $\sigma$ is the sigmoid function, $n$ is sampled only from labeled data, $l$ is sampled from both censored and uncensored data under the condition of $y^{n} < y^{l}$, and $N_c$ is the number of combinations of $\{n,l\}$ in the training set. In this experiment, we used \textbf{CTR-N} with the same implementation as CTR-N in the previous experiment. Note that we could not use CTR-D here since we needed to handle $D=37$ attributes, which would cause a combinatorial explosion for $K$, i.e., when we set the number of segments to $100$, $K$ becomes $100^{26}$.

\paragraph{Methods for comparison.}
The methods for comparison were the same as in the synthetic dataset experiments.

\paragraph{Combinations of CTR and time-series models.}
We also examined combinations of CTR-N with time-series models LSTM, Transformer, and ODERNN by concatenating their representations as inputs for $f$, i.e., $f(\bz, \bh_{\mathrm{LSTM}}(\bz_\mathrm{TS}),\bx_{\mathrm{dem}})$, $f(\bz, \bh_{\mathrm{Transformer}}(\bz_\mathrm{TS}),\bx_{\mathrm{dem}})$, and $f(\bz, \bh_{\mathrm{ODERNN}}(\bz_\mathrm{TS}),\bx_{\mathrm{dem}})$, respectively (\textbf{CTR+LSTM}, \textbf{CTR+Transformer}, and \textbf{CTR+ODERNN}).

\section{Details on Experiments with Case Study Dataset}
The case study dataset was a real-world EHR dataset provided by one of the largest hospitals in Japan that has maintained a big database of more than $400,000$ patients since $2004$~\cite{inaguma2020increasing}.
We worked with six datasets for six kinds of complications of diabetes mellitus: hyperosmolar (HYP), nephrology (NEP), retinopathy (RET), neuropathy (NEU), vascular disease (VAS), and other complications (OTH).
The numbers of samples were HYP: $N=15,428$, NEP: $N=15,862$, RET: $N=15,882$, NEU: $N=15,644$, VAS: $N=15,536$, and OTH: $N=15,591$.

The input EHR for the $n$-th patient was represented by $\{\bX^{\{\!n\!\}}, \bt^{\{\!n\!\}}\}$, where the raw observations $\bX^{\{\!n\!\}}$ were real-valued results of lab tests whose $m$-th observation was $\bx^{\{n,m\}}$, the number of attributes for $\bx^{\{n,m\}}$ was $D=26$, and the observation length for $\bX^{\{\!n\!\}}$ was $M_n$, which differed over patients. Details on the lab tests are summarized in Table~\ref{tabListLabTests}.
$\bt^{\{\!n\!\}}$ represents the corresponding timestamps. The observation intervals between them vary over time. Thus, the stay time information and the direct modeling of it as a variable is crucial for prediction with this dataset the same as in the previous experiments on synthetic data.
The MET label for the $n$-th input EHR was $y^{\{\!n\!\}} > 0$, which is the number of days over which a complication developed.
We used the patients' demographic information, such as their age and sex, which are summarized in Table~\ref{tabListDemoInfo}, with their latest lab test results as $\bx_{\mathrm{dem}}^{\{\!n\!\}}$ and input it to $f$ in addition to $\bz$ as $f(\bz, \bx_{\mathrm{dem}}^{\{\!n\!\}})$. We used their actual age, ``Age," as a natural number-valued feature and also applied one-hot encoding for each generation as a binary feature, such as ``20s (binary)." For sex, we used one-hot encoding again as ``Male (binary)" or ``Female (binary)."
All features were z-normalized with the exception of binary features.

\paragraph{Implementation.}
In this experiment, since there were censored data, we used the same loss as the Physionet experiments~\cite{liu2018early}.
In this experiment, \textbf{CTR-K} and \textbf{CTR-N} had the same implementations as CTR-K and CTR-N in the previous experiments. Note that we could not use CTR-D here since we needed to handle $D=26$ attributes, which would cause a combinatorial explosion for $K$, i.e., when we set the number of segments to $100$, $K$ becomes $100^{26}$.

\paragraph{Methods for comparison.}
The proposed method was compared with five state-of-the-art methods.
\textbf{RankSVX} is the method proposed in~\cite{liu2018early} that uses the loss in Eq.~\eqref{EqLossEHR} and has the same prediction model $f$ as our method without CTR, i.e., $f(\bx_{\mathrm{dem}})$. For its input, we used mean, standard deviation, and $\{0.1, 0.25, 0.5, 0.75, 0.9\}$ quantiles in the input time series having the stay time $\bd$ treated as another column for the input time series.
\textbf{CNN}, \textbf{GRU}, and \textbf{LSTM} have the same prediction model $f$ as our method but have outputs of CNN~\cite{cheng2016risk,makino2019artificial,phan2021deep}, GRU~\cite{chung2014empirical,tang2017memory}, and LSTM as inputs for $f$ instead of $\bz$, i.e., $f(\bh_{\mathrm{CNN}}(\bz_\mathrm{TS}),\bx_{\mathrm{dem}})$, $f(\bh_{\mathrm{GRU}}(\bz_\mathrm{TS}),\bx_{\mathrm{dem}})$, and $f(\bh_{\mathrm{LSTM}}(\bz_\mathrm{TS}),\bx_{\mathrm{dem}})$, respectively, where the number of hidden states in each model is the same as CTR-N, and $\bz_\mathrm{TS}$ is a time-series representation with stay time.
They are based on the time-series representation.

\paragraph{Combinations of CTR and time-series models.}
We also examined combinations of CTR-N with time-series models CNN, GRU, and LSTM by concatenating their representations as inputs for $f$, i.e., $f(\bz, \bh_{\mathrm{CNN}}(\bz_\mathrm{TS}),\bx_{\mathrm{dem}})$, $f(\bz, \bh_{\mathrm{GRU}}(\bz_\mathrm{TS}),\bx_{\mathrm{dem}})$, and $f(\bz, \bh_{\mathrm{LSTM}}(\bz_\mathrm{TS}),\bx_{\mathrm{dem}})$, respectively (\textbf{CTR+CNN}, \textbf{CTR+GRU}, and \textbf{CTR+LSTM}).

\begin{table}[t]
\caption{List of lab tests in EHR.}
\label{tabListLabTests}
\centering
\begin{tabular}{cc}
\toprule
 No&Lab test\\
 \midrule
 \midrule
1&Blood sugar level\\
2&BMI\\
3&BUN\\
4&CRP\\
5&Diastolic blood pressure\\
6&eGFR\\
7&Fe\\
8&Ferritin\\
9&HbA1c\\
10&HDL cholesterol\\
11&Hematocrit level\\
12&Hemoglobin\\
13&LDL cholesterol\\
14&MCH\\
15&MCHC\\
16&MCV\\
17&Serum albumin\\
18&Serum creatinine\\
19&Systolic blood pressure\\
20&Total cholesterol\\
21&Transferrin saturation\\
22&Triglyceride\\
23&UIBC\\
24&Uric acid\\
25&Urine occult blood\\
26&Urine protein\\
\bottomrule
\end{tabular}
\end{table}

\begin{table}[t]
\caption{List of demographic information of patients.}
\label{tabListDemoInfo}
\centering
\begin{tabular}{cc}
\toprule
 No&Demographic information\\
 \midrule
 \midrule
 1&Age\\
2&0s (binary)\\
3&10s (binary)\\
4&20s (binary)\\
5&30s (binary)\\
6&40s (binary)\\
7&50s (binary)\\
8&60s (binary)\\
9&70s (binary)\\
10&80s (binary)\\
11&90s (binary)\\
12&100s (binary)\\
13&Male (binary)\\
14&Female (binary)\\
\bottomrule
\end{tabular}
\end{table}

\paragraph{Results.}
The results of the mean and standard error of the C-index are listed in Table~\ref{tabDiabetesResults2}.
For most of the six tasks having over $10,000$ samples each, the performances of the methods with CTR were better than those of the methods without CTR by a sufficient margin in terms of standard error.
These results demonstrate that CTR can improve the C-index in the MET prediction problem with real-world EHR.
We found that CTR-N based on the neural network performed significantly better than CTR-K based on the kernel, which demonstrates that CTR-N learns states from data well even for real-world EHR.

CTR-N achieved the best performance on average in comparison with the single models. In addition, when looking at results for combinations of CTR-N and other models, CTR+CNN, CTR+GRU, and CTR+LSTM, even for tasks where the performances of CTR-N alone were worse than the others, we can see that adding CTR-N to them improved their performance further, such as results on RET and VAS. Also, these combinations almost constantly enhanced the performances of the original ones, which shows the high modularity of CTR to work complementarily with other models.
This shows that CTR and the time-series models captured different temporal natures.
We can determine which type of temporal natures to take into account with the training dataset by training $f$ and putting it on top of these models.

\begin{table*}[t]
\centering
\small
\begin{tabular}{ccccccc|c}
\toprule
{} & HYP & NEP & RET & NEU & VAS & OTH & Average\! \\
\hline
\!RankSVX\!     &  \!0.566$\!\pm\!$0.052\! &  \!0.688$\!\pm\!$0.008\! &  \!0.628$\!\pm\!$0.028\! &  \!0.539$\!\pm\!$0.024\! &   \!0.458$\!\pm\!$0.020\! &  \!0.684$\!\pm\!$0.019\! &   \!0.622\! \\
\!CNN\! &\!  0.608$\!\pm\!$0.033 \!&\!  0.668$\!\pm\!$0.007 \!&\!  0.697$\!\pm\!$0.012 \!&\!  0.578$\!\pm\!$0.021 \!&\!  0.469$\!\pm\!$0.015 \!&\!  0.726$\!\pm\!$0.017 \!&\!   0.651 \\
\!GRU\!       &  \!0.551$\!\pm\!$0.064\! &   \!0.705$\!\pm\!$0.010\! &  \!$\bm{0.731}$$\!\pm\!$0.009\! &  \!0.559$\!\pm\!$0.022\! &  \!$\bm{0.516}$$\!\pm\!$0.024\! &  \!0.712$\!\pm\!$0.015\! &   \!0.671\! \\
\!LSTM\!       &  \!0.589$\!\pm\!$0.026\! &  \!0.689$\!\pm\!$0.012\! &  \!0.717$\!\pm\!$0.013\! &  \!0.569$\!\pm\!$0.023\! &  \!0.503$\!\pm\!$0.035\! &  \!0.718$\!\pm\!$0.015\! &   \!0.664\! \\
\hdashline[1pt/1pt]
\!CTR-K\!     &  \!0.585$\!\pm\!$0.047\! &  \!0.688$\!\pm\!$0.015\! &   \!0.647$\!\pm\!$0.010\! &   \!0.550$\!\pm\!$0.025\! &  \!0.463$\!\pm\!$0.023\! &  \!0.704$\!\pm\!$0.021\! &   \!0.633\! \\
\!CTR-N\!     &  \!$\bm{0.612}$$\!\pm\!$0.026\! &   \!$\bm{0.739}$$\!\pm\!$0.010\! &  \!0.721$\!\pm\!$0.026\! &   \!$\bm{0.608}$$\!\pm\!$0.020\! &  \!0.481$\!\pm\!$0.013\! &   \!$\bm{0.741}$$\!\pm\!$0.020\! &   \!$\bm{0.687}$\! \\
\midrule
\!CTR+CNN\! &\!  \textbf{\textit{0.614}}$\!\pm\!$0.025 \!&\!  0.682$\!\pm\!$0.012 \!&\!  0.728$\!\pm\!$0.009 \!&\! 0.579$\!\pm\!$0.014 \!&\!  0.478$\!\pm\!$0.026 \!&\!  0.730$\!\pm\!$0.012 \!&\!   0.666 \\
\!CTR+GRU\! &  \!\textbf{\textit{0.612}}$\!\pm\!$0.024\! &  \!0.705$\!\pm\!$0.011\! &  \!\textbf{\textit{0.738}}$\!\pm\!$0.017\! &   0.592$\!\pm\!$0.020\! & \!\textbf{\textit{0.531}}$\!\pm\!$0.019\! & \!0.723$\!\pm\!$0.006\! & \!0.682\! \\
\!CTR+LSTM\! &  \!0.583$\!\pm\!$0.036\! &  \!0.708$\!\pm\!$0.007\! &  \!\textbf{\textit{0.745}}$\!\pm\!$0.008\! &     \!0.600$\!\pm\!$0.020\! &  \!\textbf{\textit{0.534}}$\!\pm\!$0.022\! &  \!0.734$\!\pm\!$0.011\! &   \!\textbf{\textit{0.688}}\! \\
\bottomrule
\end{tabular}
\caption{Comparison of C-index on real EHR data (higher is better). Upper seven results show comparison among single models, and best results are in bold. Bottom three results are for combinations of CTR and other models, and results further improved from best results in single models are in bold italic. Confidence intervals are standard errors.}
\label{tabDiabetesResults2}
\end{table*}

\begin{figure}[t]
    \centering
    \includegraphics[width=84mm]{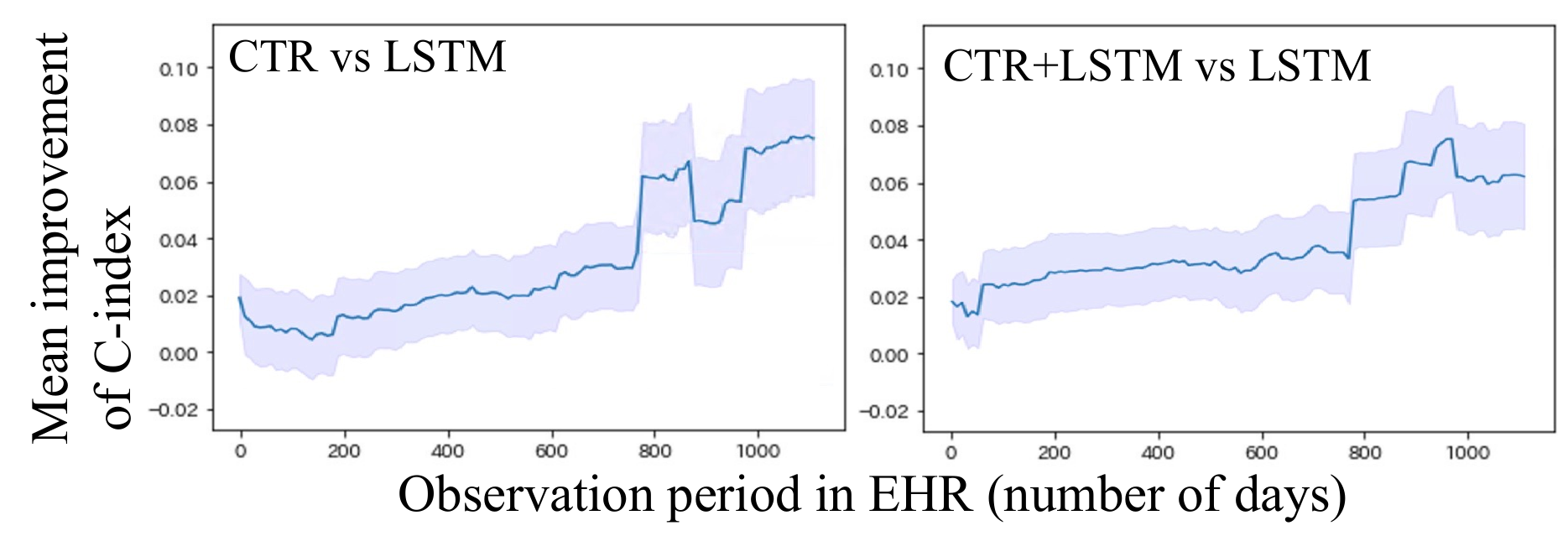}
    \caption{Analysis of performance improvement by CTR compared with \emph{LSTM} over different observation periods (higher is better).}
    \label{FigExDiffPeriod2}
\end{figure}

\begin{figure}[t]
    \centering
    \includegraphics[width=84mm]{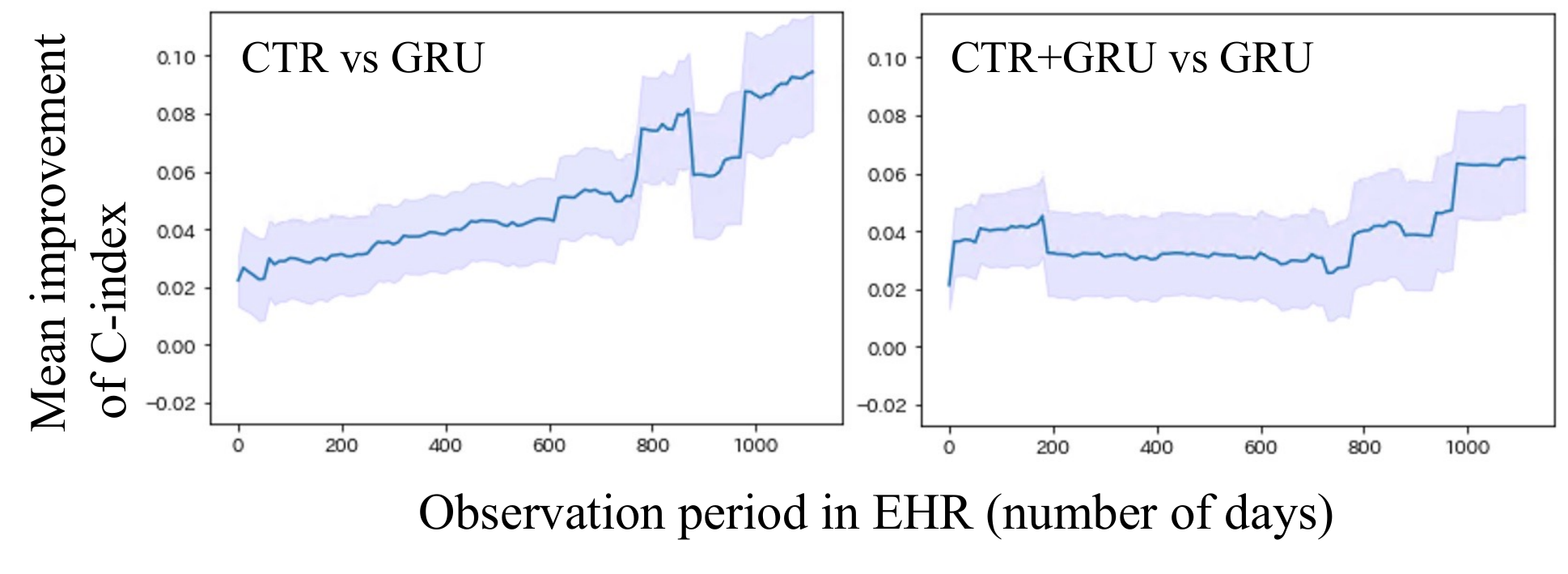}
    \caption{Analysis of performance improvement by CTR compared with \emph{GRU} over different observation periods (higher is better).}
    \label{FigExDiffPeriod_GRU}
\end{figure}

\begin{figure}[t]
    \centering
    \includegraphics[width=84mm]{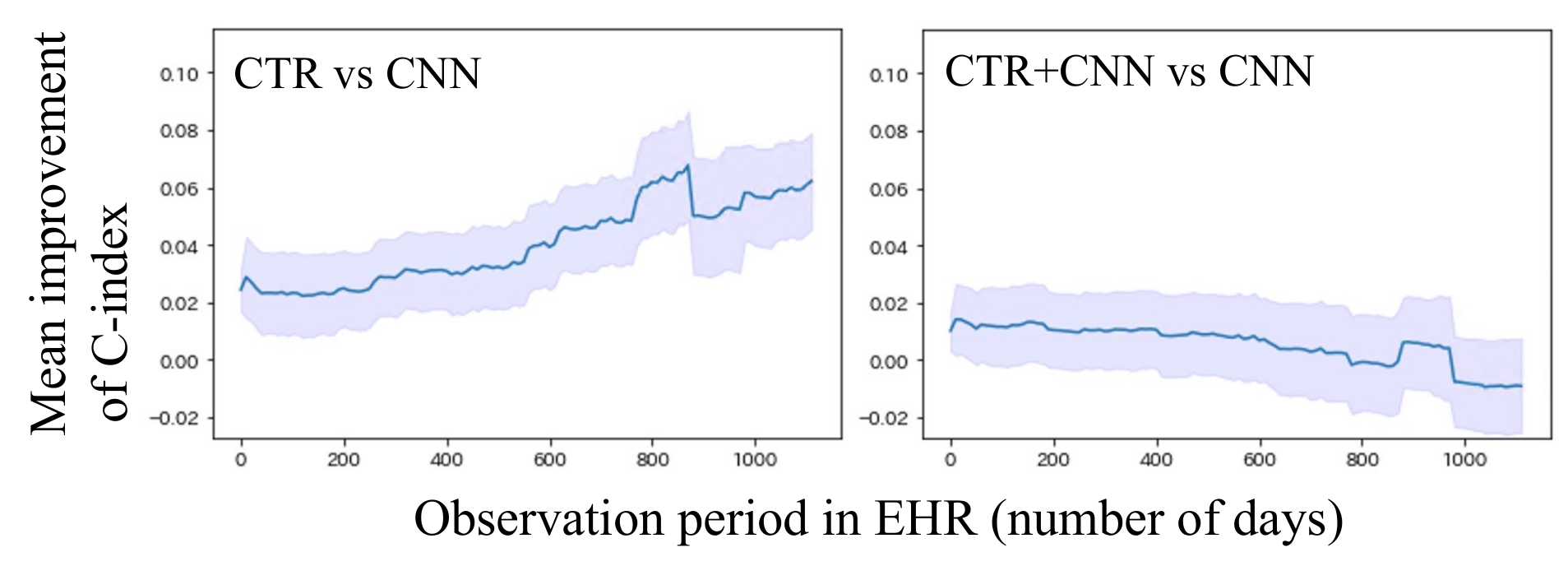}
    \caption{Analysis of performance improvement by CTR compared with \emph{CNN} over different observation periods (higher is better).}
    \label{FigExDiffPeriod_CNN}
\end{figure}

\paragraph{Performance analysis on different observation periods with LSTM, GRU, and CNN.}
Over different observation periods, we analyzed the performance improvements of CTR and CTR+LSTM compared with LSTM and the same comparison with GRU and CNN.
We plotted the mean improvements in the C-index by CTR for data having different minimum observation periods, as shown in Figs.~\ref{FigExDiffPeriod2}--\ref{FigExDiffPeriod_CNN}, where the confidence intervals are standard errors of the improvements.
The right region in the figure shows the results for data with longer observation periods.
These results demonstrate that CTR could improve the performance, especially for data with relatively longer observation periods, where cumulative health conditions are more crucial for MET prediction.


\section{Ethical/Societal Impact}
The common concern when learning from data that is collected through experiments conducted with human participants, including healthcare and medical applications, is producing estimation models biased towards or against specific groups in a population.
Recent works on fairness in machine learning~\cite{pedreshi2008discrimination,kilbertus2017avoiding,kusner2017counterfactual,nabi2018fair,lipton2018does,zhang2018equality,locatello2019fairness,singh2019policy,bera2019fair,ding2020differentially,yan2020fairness,narasimhan2020pairwise,rezaei2020fairness} are one example of help for this, and developing efficient ways of applying them to our approach would be an interesting and useful next step of our study.

Another risk of estimation that may affect human decisions would be false alerts/reports and overlooking important events.
We believe that the estimation should be carefully used as just one source of information, and it is better that actual decision-making based on this estimation is done from a broader perspective.

\end{document}